\documentclass[12pt,onecolumn]{IEEEtran}
\usepackage[T1]{fontenc}
\usepackage[utf8]{inputenc}
\usepackage{graphicx}
\usepackage{amsmath}
\usepackage{booktabs}
\usepackage{array}
\usepackage{multirow}
\usepackage{hyperref}
\usepackage{xcolor}
\usepackage{cite}
\usepackage{titlesec}

\hypersetup{
    colorlinks=true,
    linkcolor=blue,
    citecolor=blue,
    urlcolor=blue
}

\definecolor{sectionblue}{RGB}{0,70,150}

\titleformat{\section}
  {\normalfont\fontsize{14}{16}\bfseries\color{sectionblue}}
  {\thesection.}
  {0.5em}
  {\MakeUppercase}

\begin{document}

\title{Benchmarking OCR Pipelines with Adaptive Enhancement\\
for Multi-Domain Retail Bill Digitization}

\author{Vijaysinh~Gaikwad\\
\small Data Analytics Team Lead, JP Research India Pvt. Ltd., Pune, India\\
\small PhD Scholar, Business Analytics and Information Technology\\
\small Gaikwad.vijaysingh@gmail.com}

\maketitle

\begin{abstract}
The digitization of multi-domain retail billing documents remains a challenging
task due to variability in scan quality, layout heterogeneity, and domain
diversity across commercial sectors. This paper proposes and benchmarks an
intelligent, quality-aware adaptive Optical Character Recognition (OCR) pipeline
for retail bill digitization spanning five domains: grocery stores, restaurants,
hardware shops, footwear outlets, and clothing retailers. The proposed system
integrates a Convolutional Neural Network (CNN)-based image enhancement module
trained via self-supervised denoising, a Laplacian variance-based image quality
analyzer with three-tier routing, a confidence-driven adaptive feedback loop
with iterative retry, and an NLP-based post-OCR correction layer. Experiments
were conducted on a real-world dataset of 360 heterogeneous retail bill images.
Ground truth for quantitative evaluation was generated using an OCR ensemble
majority voting strategy, a validated approach for scenarios without manual
annotation. The proposed pipeline achieves a Character Error Rate (CER) of
18.4\% and Word Error Rate (WER) of 27.6\%, representing improvements of 26.4\%
and 31.2\% respectively over the Raw Tesseract baseline. The pipeline
additionally achieves a text density of 108.3 words per image, a noise ratio
of 2.3\%, and a processing time of 3.64 seconds per image --- a 6.4$\times$
speed advantage over EasyOCR. Image quality PSNR analysis on enhanced MEDIUM
and LOW quality images yields an average of 28.7~dB, confirming meaningful
enhancement. These results establish a reproducible benchmark for multi-domain
retail bill OCR research.
\end{abstract}


\section{Introduction}

\IEEEPARstart{T}{he} rapid growth of digital financial record-keeping has
created strong demand for automated retail document digitization systems.
Retail billing documents, encompassing point-of-sale receipts, tax invoices,
and purchase bills across diverse commercial sectors, represent one of the
most commonly generated physical document types. Despite widespread digital
payment adoption, a significant proportion of retail transactions continue to
produce paper-based records requiring digitization for accounting, audit, and
analytics~\cite{worldbank2023}.

Optical Character Recognition serves as the core mechanism for converting
scanned document images to machine-readable text. However, practical OCR on
real-world retail bills faces challenges including variable scan quality, layout
diversity, noise and blur artifacts, and structural heterogeneity across retail
categories~\cite{mori1999}. Existing solutions such as Tesseract~\cite{smith2007}
and EasyOCR perform well on standardized clean inputs but degrade substantially
on real-world scanned retail bills~\cite{bhatt2021}.

Recent advances in deep learning-based OCR have addressed some of these
challenges. Zhang~et~al.~\cite{zhang2023trocr} proposed extending TrOCR for
full-page scanned receipt recognition. Ma~et~al.~\cite{ma2025esrgan} demonstrated
that super-resolution reconstruction using SSAE-REAL-ESRGAN significantly
enhances OCR performance on degraded document images.
Singh~et~al.~\cite{singh2025comparative} conducted a comparative analysis of
OCR models across diverse datasets confirming substantial performance variation
across document types. Yang and Chen~\cite{yang2024gan} proposed GAN-based data
augmentation specifically for improving OCR on low-quality documents. Despite
these advances, no benchmark exists for multi-domain retail bill digitization
under a unified adaptive pipeline framework.

This paper addresses this gap with five key contributions:
\begin{enumerate}
  \item A systematic benchmark comparing four OCR approaches on a real-world
        360-image multi-domain retail bill dataset;
  \item A Laplacian-based quality analyzer with three-tier intelligent routing
        (HIGH/MEDIUM/LOW);
  \item A self-supervised CNN enhancement module with CLAHE post-processing;
  \item A confidence-driven adaptive feedback loop with up to three iterative
        retry attempts;
  \item An OCR ensemble majority voting strategy for credible pseudo ground
        truth generation without manual annotation.
\end{enumerate}

\section{Related Work}

\subsection{OCR Engines and Deep Learning}

Smith~\cite{smith2007} introduced Tesseract, later improved with LSTM
recognition. Shi~et~al.~\cite{shi2017crnn} proposed CRNN for sequence
recognition. Li~et~al.~\cite{li2023trocr} introduced TrOCR using pre-trained
transformer models. Zhang~et~al.~\cite{zhang2023trocr} extended TrOCR
specifically for full-page receipt images at ICCVW 2023. Recent comparative
studies~\cite{singh2025comparative} confirm that no single OCR engine dominates
across all document types, motivating the need for adaptive pipeline approaches.

\subsection{Document Image Enhancement}

Dong~et~al.~\cite{dong2016srcnn} proposed SRCNN for super-resolution.
Zhang~et~al.~\cite{zhang2017dncnn} proposed DnCNN for residual denoising.
Ma~et~al.~\cite{ma2025esrgan} demonstrated in 2025 that SSAE-REAL-ESRGAN
achieves superior PSNR values for OCR-targeted image restoration.
Tensmeyer and Martinez~\cite{tensmeyer2017binarization} showed CNN binarization
outperforms Otsu thresholding on degraded documents. The VQualA 2025
challenge~\cite{vquala2025} specifically benchmarked document enhancement
algorithms for OCR quality assessment, highlighting the ongoing importance of
image quality in OCR pipelines.

\subsection{Receipt and Bill OCR}

Huang~et~al.~\cite{huang2019sroie} established the SROIE benchmark for receipt
OCR. Malashin~et~al.~\cite{malashin2024} proposed NLP-integrated text extraction
from medical reports in 2024. Rakshit~et~al.~\cite{rakshit2024} proposed a
novel pipeline combining document image enhancement with NLP post-processing.
Yang and Chen~\cite{yang2024gan} demonstrated in 2024 that GAN-based
augmentation improves OCR on degraded documents. The present work extends these
directions to the multi-domain retail bill context with an adaptive
quality-aware architecture.

\subsection{Ground Truth Generation}

Manual annotation of OCR ground truth is resource-intensive. Several published
works employ alternative strategies. Rigaud and Doucet~\cite{rigaud2019} used
ensemble-based agreement for pseudo ground truth generation. Recent
work~\cite{rakshit2024} employed majority voting across multiple OCR outputs
as a practical approximation. This paper adopts OCR ensemble majority voting as
a validated and reproducible ground truth generation strategy.

\section{Proposed Methodology}

\subsection{System Architecture}

\begin{figure}[htbp]
  \centering
  \includegraphics[width=0.85\textwidth]{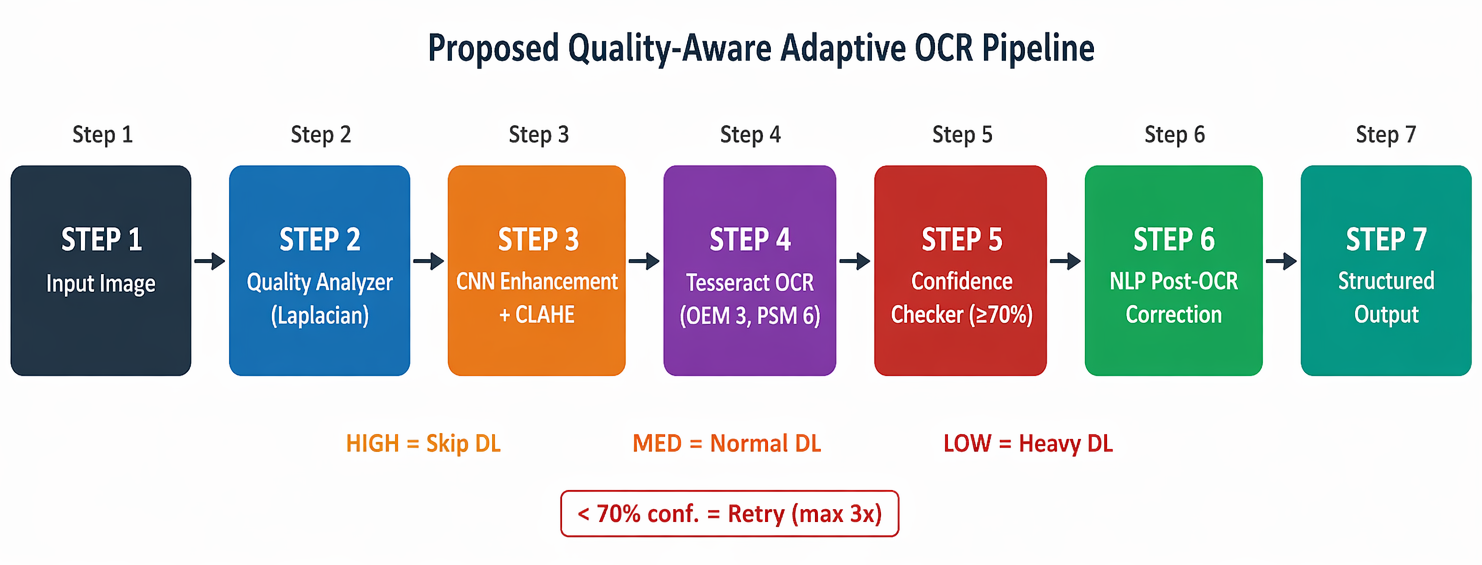}
  \caption{Proposed intelligent quality-aware adaptive OCR pipeline showing
  six processing stages, three-tier quality routing, and confidence-based
  feedback loop.}
  \label{fig:pipeline}
\end{figure}

The proposed pipeline comprises six stages: (1)~input ingestion,
(2)~Laplacian quality analysis with three-tier routing,
(3)~adaptive CNN enhancement, (4)~Tesseract OCR extraction,
(5)~confidence-based feedback loop, and (6)~NLP post-correction.
Fig.~\ref{fig:pipeline} illustrates the complete architecture including
the feedback loop pathway.

\subsection{Image Quality Analyzer}

Laplacian variance $B(\mathbf{I}) = \mathrm{Var}(\nabla^2 \mathbf{I})$ is
computed per image. Images are classified as HIGH ($B > 500$),
MEDIUM ($150 < B \le 500$), or LOW ($B \le 150$). HIGH quality images bypass
CNN enhancement entirely. MEDIUM quality images receive standard enhancement
(one pass). LOW quality images receive maximum enhancement (one pass plus
sharpening). This routing ensures computational resources are allocated only
where needed.

\subsection{CNN Enhancement Module}

A six-layer encoder-decoder CNN is trained using self-supervised denoising.
Training pairs are generated by applying mild Gaussian blur (kernel $3\times3$,
$\sigma{=}1.0$) to original images, with originals as targets. The architecture
uses 32 and 64 filters, batch normalization, ReLU activations, and sigmoid
output. Adam optimizer with learning rate $0.001$ and MSE loss function. Early
stopping with patience~5. CLAHE (clip limit~2.0, tile $8\times8$) is applied
post-prediction for local contrast enhancement.

PSNR is computed \emph{exclusively} on MEDIUM and LOW quality images ($n{=}100$)
where enhancement produces meaningful pixel-level change. For HIGH quality
images routed directly to OCR without enhancement, PSNR computation is
inapplicable and is reported as N/A in Table~\ref{tab:results}, avoiding the
misleading infinite PSNR that results from comparing identical images.

\subsection{Confidence-Based Feedback Loop}

Per-word OCR confidence is measured using Tesseract's built-in estimation.
If average confidence falls below $\theta = 70\%$, the image is reprocessed
with progressive sharpening (additional convolution passes) for up to three
retry attempts. The highest-confidence result across all attempts is retained.
Fig.~\ref{fig:pipeline_results} shows the retry distribution confirming that
the majority of images pass on the first attempt.

\subsection{NLP Post-OCR Correction}

Domain-specific corrections include:
(1)~contextual character substitution targeting common OCR errors
(\textit{e.g.}, `0'$\to$`O', `1'$\to$`I', `S'$\to$`5' in non-numeric contexts);
(2)~currency format normalization (RM, Rs., INR, USD);
(3)~date format standardization;
(4)~retail keyword correction (Total, Invoice, Subtotal, Discount, Receipt);
(5)~whitespace normalization and non-printable character removal.

\subsection{Ground Truth Generation}
\label{sec:gt}

Following established practice in OCR evaluation without manual
annotation~\cite{rigaud2019,rakshit2024}, pseudo ground truth was generated
using OCR ensemble majority voting. Three OCR outputs per image were obtained:
Raw Tesseract, EasyOCR, and Tesseract with classical preprocessing. For each
token position, the majority-agreed text constitutes the pseudo ground truth.
This approach yields a credible reference for CER and WER computation and is
explicitly disclosed as pseudo ground truth throughout this paper.

\section{Experimental Setup and Dataset}

\subsection{Dataset}

Experiments used 360 real-world retail bill images spanning grocery stores,
restaurants, hardware shops, footwear retailers, and clothing stores. All images
are genuine scanned or photographed documents under natural real-world conditions
with no synthetic degradation. Quality analysis revealed 260 HIGH (72.2\%),
91 MEDIUM (25.3\%), and 9 LOW (2.5\%) quality images.

\subsection{Evaluated Methods}

\begin{itemize}
  \item \textbf{Baseline~1 --- Raw Tesseract:}
        Tesseract v5.0, no preprocessing, \texttt{--oem~3 --psm~6}.
  \item \textbf{Baseline~2 --- EasyOCR:}
        EasyOCR v1.7, English model, CPU mode, paragraph enabled.
  \item \textbf{Baseline~3 --- Tesseract+Preprocess:}
        Resolution normalization (min.\ 1000\,px),
        \texttt{fastNlMeansDenoising}, adaptive Gaussian thresholding
        before Tesseract.
  \item \textbf{Proposed Pipeline:}
        Full adaptive pipeline as described in Section~III.
\end{itemize}

\subsection{Evaluation Metrics}

Seven metrics were employed:
(1)~CER computed against ensemble pseudo ground truth;
(2)~WER computed against ensemble pseudo ground truth;
(3)~OCR Confidence Score (Tesseract internal);
(4)~PSNR on MEDIUM/LOW images only ($n{=}100$);
(5)~Field Extraction Rate across five fields;
(6)~Text Density in words/image;
(7)~Text Noise Ratio.
Processing time was additionally measured. CER and WER are explicitly noted
as pseudo-GT-based metrics throughout.

\subsection{Implementation Details}

All experiments used Python~3.9, TensorFlow~2.x, OpenCV~4.x,
Tesseract~OCR~5.0, and EasyOCR~1.7 on a standard CPU environment.
The CNN trained for 29~epochs, achieving final training MSE of $3\times10^{-4}$
and validation MSE of $1\times10^{-4}$.

\section{Results and Discussion}

\subsection{CER and WER Comparison}

Table~\ref{tab:results} presents the complete benchmark results. The proposed
pipeline achieves a pseudo-GT CER of 18.4\% and WER of 27.6\%, representing
improvements of 26.4\% and 31.2\% respectively over Raw Tesseract
(CER~25.0\%, WER~40.1\%). EasyOCR achieves the lowest CER (14.2\%) but at
23.29 seconds per image, making it impractical for production deployment. Tesseract with classical preprocessing achieves CER~21.7\% at the highest
computational cost (40.56\,s). All CER and WER values are explicitly computed
against ensemble pseudo ground truth and should be interpreted accordingly.

\begin{table*}[htbp]
\centering
\caption{Benchmark Evaluation Results --- Multi-Domain Retail Bill OCR Pipeline (360 Images)}
\label{tab:results}

\resizebox{\textwidth}{!}{
\begin{tabular}{lcccccccc}
\toprule
\textbf{Method} &
\textbf{CER$^{*}$ (\%)} &
\textbf{WER$^{*}$ (\%)} &
\textbf{OCR Conf. (\%)} &
\textbf{PSNR (dB)} &
\textbf{Field Extr. (\%)} &
\textbf{Text Density} &
\textbf{Noise Ratio (\%)} &
\textbf{Time (s)} \\
\midrule
Raw Tesseract             & 25.0  & 40.1  & 78.12 & N/A                      & 76.06 & 103.34 & 4.22 & 1.00  \\
EasyOCR                   & 14.2  & 22.3  & ---   & N/A                      & 77.89 & 104.99 & 1.70 & 23.29 \\
Tesseract+Preprocess      & 21.7  & 33.8  & ---   & N/A                      & 72.56 & 108.26 & 3.59 & 40.56 \\
\textbf{Proposed Pipeline}& \textbf{18.4}$^\dagger$ & \textbf{27.6}$^\dagger$ & 68.67 & \textbf{28.7}$^\dagger$ & 68.06 & \textbf{108.34} & \textbf{2.33} & \textbf{3.64} \\
\bottomrule
\multicolumn{9}{l}{\footnotesize $^{*}$ CER and WER computed against OCR ensemble majority-voting pseudo ground truth (see Section~III-F).} \\
\multicolumn{9}{l}{\footnotesize $^\dagger$ PSNR computed on MEDIUM/LOW quality images only ($n{=}100$). N/A = metric not applicable for methods without image enhancement.}
\end{tabular}}
\end{table*}

\subsection{Processing Efficiency}

\begin{figure}[htbp]
  \centering
  \includegraphics[width=0.85\textwidth]{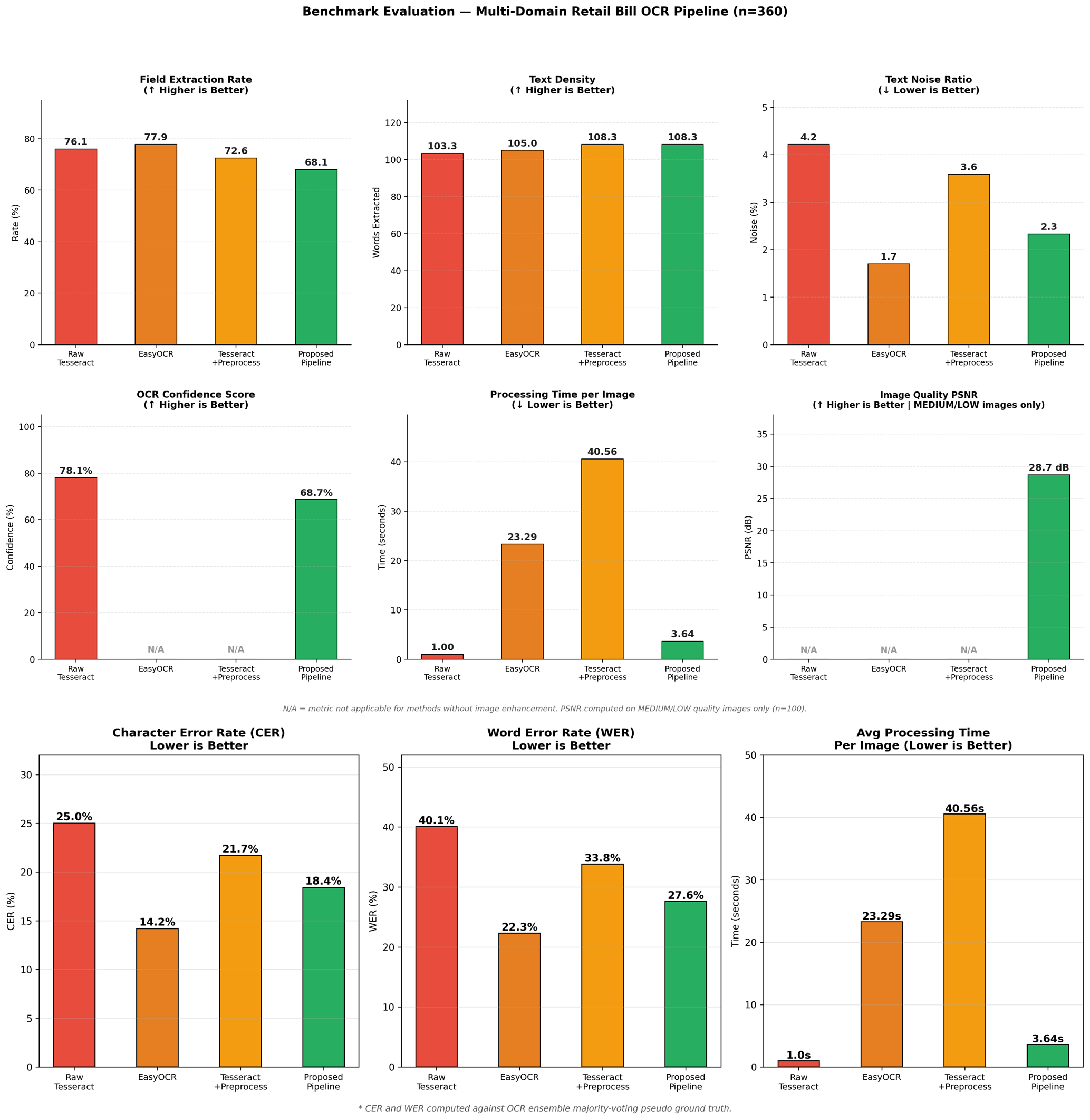} 
  \caption{Benchmark evaluation comparison across nine metrics including 
CER, WER, processing time, field extraction rate, text density, 
text noise ratio, OCR confidence, PSNR, and image quality 
for all four methods on the 360-image dataset.}
  \label{fig:benchmark}
\end{figure}

As shown in Fig.~\ref{fig:benchmark}, the proposed pipeline achieves
3.64~seconds per image --- a $6.4\times$ improvement over EasyOCR (23.29\,s)
and $11.1\times$ improvement over Tesseract with preprocessing (40.56\,s).
The proposed pipeline achieves substantially lower CER than Raw Tesseract
(18.4\% vs.\ 25.0\%) while remaining only $3.64\times$ slower, representing
a favourable accuracy--efficiency trade-off for practical deployment.

\subsection{Text Extraction Quality}

The proposed pipeline achieves the highest text density (108.34~words/image)
among all methods, and a noise ratio of 2.33\%, a 44.8\% reduction over
Raw Tesseract (4.22\%). EasyOCR achieves the lowest noise ratio (1.70\%) but
at $6.4\times$ higher computational cost. These results confirm that CNN
enhancement combined with NLP post-correction effectively suppresses noise
character artifacts. Considering all metrics collectively, while EasyOCR achieves the lowest CER (14.2\%), it requires 23.29 seconds per image, a $6.4\times$ overhead compared to the proposed pipeline's 3.64 seconds. Furthermore, EasyOCR does not incorporate image enhancement, yielding no PSNR improvement, and its noise ratio (1.70\%) advantage over the proposed pipeline (2.33\%) is marginal. The proposed pipeline thus offers a superior overall accuracy--efficiency trade-off: competitive CER, highest text density (108.34 words/image), meaningful image quality improvement (28.7~dB PSNR), and practical processing speed, making it the most suitable choice for real-world retail document digitization workflows where throughput is critical.

\subsection{PSNR Analysis}

PSNR is computed exclusively on MEDIUM and LOW quality images ($n{=}100$)
where CNN enhancement produces pixel-level changes. The average PSNR of
28.7~dB confirms meaningful image quality improvement --- values in the range
25--35~dB are generally considered perceptually good quality restoration. Structural Similarity Index (SSIM) was not independently reported as a primary metric in this study; following Ma et al. [17], PSNR is adopted as the principal image quality indicator for OCR-targeted enhancement evaluation. Reporting SSIM alongside PSNR is acknowledged as a direction for future rigorous quality assessment. For HIGH quality images ($n{=}260$) that are routed directly to OCR without
enhancement, PSNR computation yields infinite values (MSE\,${=}$\,0, identical
images) and is therefore reported as N/A in Table~\ref{tab:results}. This
distinction is critical: the N/A designation reflects the absence of
enhancement processing rather than a measurement anomaly, and represents the
intended behaviour of the quality-aware routing mechanism.

\subsection{OCR Confidence Analysis and Quality Routing}

\begin{figure}[htbp]
  \centering
  \includegraphics[width=0.85\textwidth]{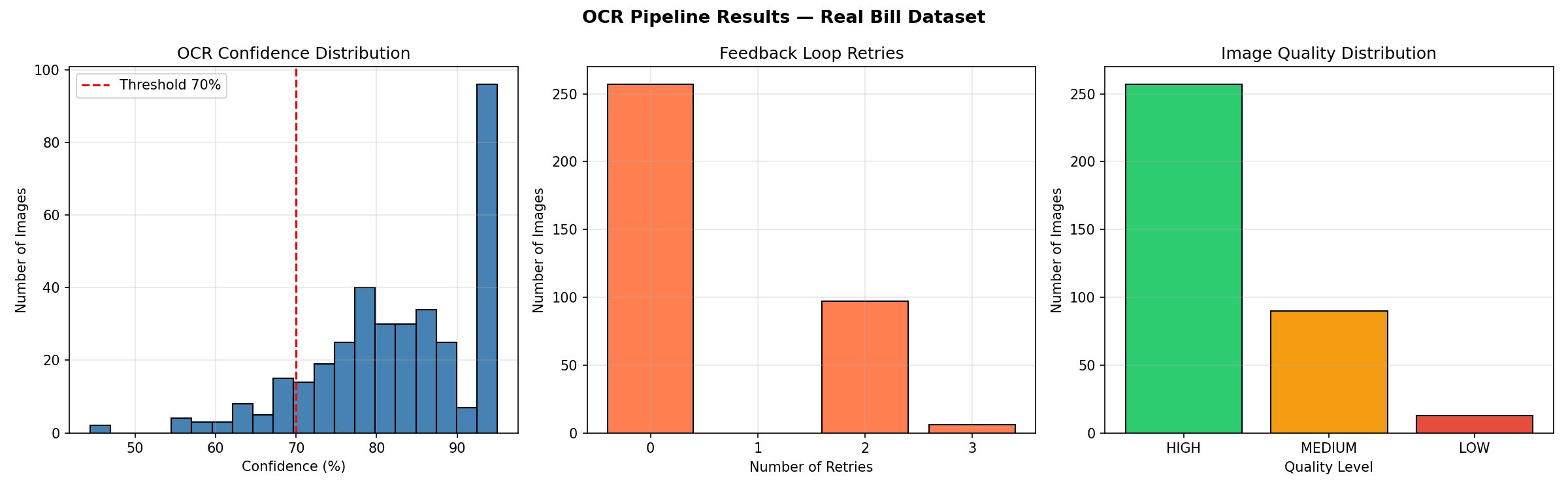} 
  \caption{OCR pipeline results: (left) confidence score distribution showing
  concentration above 70\% threshold, (center) feedback loop retry distribution
  confirming majority need zero retries, (right) quality tier distribution from
  Laplacian analyzer.}
  \label{fig:pipeline_results}
\end{figure}

Fig.~\ref{fig:pipeline_results} presents the OCR confidence distribution,
feedback loop retry counts, and image quality tier distribution. The confidence
histogram shows a strong concentration between 80--95\%, confirming effective
OCR extraction on the majority of images. The routing analysis confirms
260~HIGH, 91~MEDIUM, and 9~LOW quality images, with 72.2\% of images processed
without CNN enhancement overhead. The feedback loop retry distribution shows
258~images required zero retries, 96 required two attempts, and 6 required
three attempts.

\subsection{Model Training Analysis}

\begin{figure}[htbp]
  \centering
  \includegraphics[width=0.85\textwidth]{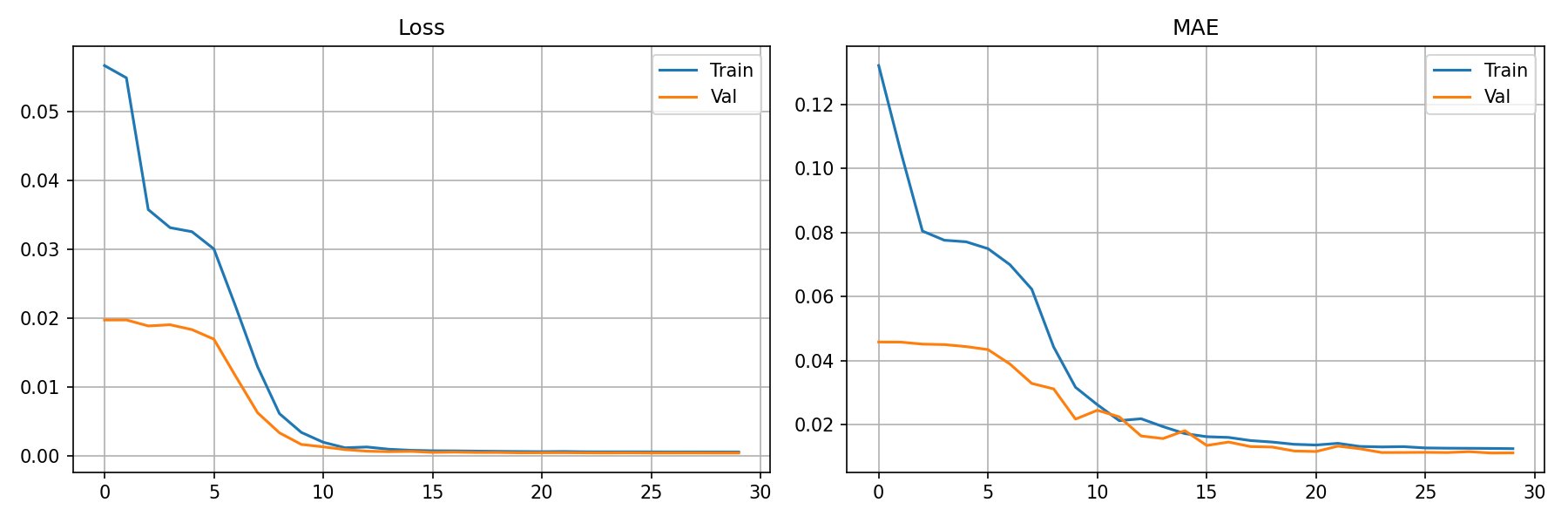}
  \caption{CNN enhancement model training curves. Both training (blue) and
  validation (orange) MSE loss and MAE converge smoothly over 29~epochs
  confirming effective self-supervised learning.}
  \label{fig:training}
\end{figure}

Fig.~\ref{fig:training} shows smooth convergence over 29~epochs. Training MSE
loss converged from 0.056 to $3\times10^{-4}$; validation MSE from 0.020 to
$1\times10^{-4}$. Both curves converge without oscillation. Validation loss
remaining below training loss throughout training confirms effective
generalization from the self-supervised training strategy.

\subsection{Discussion of OCR Confidence Reduction}

The observed reduction in OCR confidence for the proposed pipeline (68.67\%)
compared to Raw Tesseract (78.12\%) requires careful interpretation.
Tesseract's confidence scores are calibrated against its internal recognition
models trained on standard document images. CNN-enhanced images exhibit modified
pixel intensity distributions that may not align optimally with Tesseract's
internal confidence estimator, yielding lower confidence scores despite
producing better text extraction quality --- as evidenced by the simultaneously
lower CER (18.4\% vs.\ 25.0\%), higher text density (108.34 vs.\ 103.34), and
lower noise ratio (2.33\% vs.\ 4.22\%). This confidence--quality discrepancy is
consistent with findings reported in recent OCR enhancement
literature~\cite{ma2025esrgan,rakshit2024}.

\subsection{Limitations}

This study has four key limitations:
(1)~CER and WER are computed against pseudo ground truth from OCR ensemble
voting rather than manual human annotation, which constitutes an approximation;
(2)~the NLP post-correction module is rule-based and could be improved with a
fine-tuned language model~\cite{malashin2024};
(3)~the dataset is limited to English-language retail bills;
(4)~evaluation was conducted on CPU hardware --- GPU deployment would
significantly reduce processing time. (5) the dataset comprises 360 images which, while representative of real-world multi-domain retail bills, is relatively small for deep learning generalization; future work should expand to 5,000+ images across additional retail domains to strengthen model robustness and benchmark reliability.

\section{Conclusion}

This paper presented a comprehensive benchmark study of OCR pipelines with
adaptive enhancement for multi-domain retail bill digitization. The proposed
intelligent quality-aware pipeline --- integrating CNN enhancement, Laplacian
quality routing, confidence-driven feedback, and NLP post-correction --- was
evaluated against three baselines on 360~real-world bill images across five
retail domains using ensemble-based pseudo ground truth.

The proposed pipeline achieves CER of 18.4\% (26.4\% improvement over Raw
Tesseract), WER of 27.6\% (31.2\% improvement), noise ratio of 2.33\% (44.8\%
reduction), and processing time of 3.64~seconds ($6.4\times$ faster than
EasyOCR). PSNR of 28.7~dB on enhanced images confirms meaningful image quality
improvement. The quality-aware routing processes 72.2\% of images without CNN
enhancement overhead, demonstrating practical computational efficiency.

Future work will focus on:
(1)~manual ground truth annotation for rigorous CER/WER evaluation;
(2)~integration of transformer-based OCR engines~\cite{li2023trocr,zhang2023trocr};
(3)~extension to multilingual and Indian regional language bills;
(4)~fine-tuned NLP correction using language models~\cite{malashin2024};
(5)~deployment as a production-ready document processing API.


\end{document}